\documentclass[conference]{IEEEtran}

\usepackage{cite}
\usepackage{amsmath,amssymb,amsfonts}
\usepackage{algorithmic}
\usepackage{algorithm}
\usepackage{graphicx}
\usepackage{textcomp}
\usepackage{xcolor}
\usepackage{url}
\usepackage{hyperref}
\usepackage{booktabs}
\usepackage{subcaption} 
\usepackage{multirow}
\usepackage{enumitem}
\usepackage{tabularx}
\usepackage{array}
\usepackage[section]{placeins}
\usepackage{arydshln}
\usepackage{ragged2e}

\newcolumntype{Y}{>{\raggedright\arraybackslash}X}
\newcolumntype{A}{>{\hsize=0.2\hsize\raggedright\arraybackslash}X}
\newcolumntype{B}{>{\hsize=0.8\hsize\raggedright\arraybackslash}X}
\newcolumntype{S}{>{\RaggedRight\arraybackslash}p{0.12\columnwidth}}
\newcolumntype{L}{>{\RaggedRight\arraybackslash}p{0.25\columnwidth}}
\newcolumntype{D}{>{\RaggedRight\arraybackslash}p{0.53\columnwidth}}
\hypersetup{colorlinks=true,citecolor=blue,linkcolor=black,urlcolor=blue}

\def\BibTeX{{\rm B\kern-.05em{\sc i\kern-.025em b}\kern-.08em
    T\kern-.1667em\lower.7ex\hbox{E}\kern-.125emX}}

\makeatletter
\newcommand{\pcite}[1]{\@ifundefined{b@#1}{}{\,\textsuperscript{\citenum{#1}}}}
\makeatother

\newcommand{\ieeecopyrightnotice}{%
  \footnotesize
  \noindent © 2025 IEEE. Personal use of this material is permitted. Permission from IEEE must be obtained
  for all other uses, in any current or future media, including reprinting/republishing this material
  for advertising or promotional purposes, creating new collective works, for resale or redistribution
  to servers or lists, or reuse of any copyrighted component of this work in other works.%
}

\def\arxivcover{}  

\begin{document}

\ifdefined\arxivcover
  \begin{titlepage}
    \thispagestyle{empty}
    \vspace*{0.28\textheight}
    \begin{center}
      \begin{minipage}{0.82\textwidth}
        \ieeecopyrightnotice

        \medskip
        {\footnotesize
        This is the accepted author manuscript for the 2025 IEEE International Conference on Data Mining Workshops (ICDMW).
        When available, please cite the version of record (DOI) and see the IEEE Xplore page.
        }
      \end{minipage}
    \end{center}
    \vfill
  \end{titlepage}
\fi

\title{SoK: Measuring What Matters for Closed-Loop Security Agents}

\author{
  \IEEEauthorblockN{Mudita Khurana}
  \IEEEauthorblockA{Application Security, Airbnb\\
  San Francisco, CA, USA\\
  muditak@airbnb.com}
  \and
  \IEEEauthorblockN{Raunak Jain}
  \IEEEauthorblockA{AI Science, Intuit\\
  Mountain View, CA, USA\\
  raunak\_jain1@intuit.com}
}

\maketitle

\ifdefined\arxivcover \setcounter{page}{1}\fi

\begin{abstract}
    Cybersecurity is a relentless arms race, with AI driven offensive systems evolving faster than traditional defenses can adapt. Research and tooling remain fragmented across isolated defensive functions, creating blind spots that adversaries exploit. Autonomous agents capable of integrating, exploit confirmation, remediation, and validation into a single closed loop offer promise, but the field lacks three essentials: a framework defining the agentic capabilities of security systems across security life cycle, a principled method for evaluating closed loop agents, and a benchmark for measuring their performance in practice. We introduce CLASP—the Closed-Loop Autonomous Security Performance framework which aligns the security lifecycle (reconnaissance, exploitation, root cause analysis, patch synthesis, validation) with core agentic capabilities (planning, tool use, memory, reasoning, reflection \& perception) providing a common vocabulary and rubric for assessing agentic capabilities in security tasks. By applying CLASP to 21 representative works, we map where systems demonstrate strengths, and where capability gaps persist. We then define the Closed-Loop Capability (CLC) Score, a composite metric quantifying both degree of loop closure and operational effectiveness, and outline the requirements for a closed loop benchmark. Together, CLASP and the CLC Score, provide the vocabulary, diagnostics, and measurements needed to advance both function level performance and measure closed loop security agents. 
\end{abstract}

\begin{IEEEkeywords}
LLM agents, autonomous pentesting, knowledge graphs, reconnaissance, patching, root cause analysis, security, security agents, closed-loop security, systematization of knowledge
\end{IEEEkeywords}

\section{Introduction}

The cybersecurity landscape is undergoing a fundamental transformation driven by the rapid adoption of Large Language Models (LLMs) agents in offensive operations. State-affiliated and criminal actors are increasingly leveraging AI to accelerate reconnaissance, social engineering, and vulnerability research~\cite{openai_microsoft_disruption_2024}. The UK's NCSC projects that AI will lower barriers and boost attack cadence through at least 2027~\cite{ncsc_ai_threat_2025}, while Microsoft Threat Intelligence reports ongoing misuse of LLMs for cyberattacks~\cite{msft_staying_ahead_2024}. This escalating threat environment, combined with increasing incident dwell times ~\cite{mandiant2024mtrends} and breach costs ~\cite{ibm2024cost}, demands for defensive actors to also operate at similar speed with autonomous capabilities.

While mostly considered an attacker function, reconnaissance and exploitation activities are useful for enterprises to understand their own gaps and thus improve their defenses~\cite{cset-vulnlifecycle}. Recent academic advances show increasing autonomous capabilities within individual offensive and defensive functions, like autonomous penetration testing (e.g., VulnBot~\cite{VulnBot_2501_13411}, Pentest-R1~\cite{PentestR1_2508_07382}), automated vulnerability remediation (e.g., APPATCH~\cite{APPATCH_2025}, PatchPilot ~\cite{PatchPilot_ICML2025}), and multi-agent root-cause analysis for microservices (Flow-of-Action~\cite{FlowOfAction_2502_08224}). 
On the other hand, industry sees values in integrated, closed-loop operationalization of security lifecycles \& operations (e.g., Google's Autonomic Security Operations, Gartner's CTEM cycles, and Microsoft's Automatic Attack Disruption).~\cite{googleASOwhitepaper2021,gartnerCTEM2023,microsoftAADdocs2025}, which allows them to reduce mean time to remediate and lower operational risk. However, these pipelines remain automated, not agentic. 

On the academic front, there has not been much push towards a closed loop integrated agentic systems. DARPA's recent AI Cyber Challenge (AIxCC) shows what end-to-end closed loop scoring for agents can look like~\cite{aixcc_scoring_2025}. The next step is making that score useful for science and engineering. Final outcomes alone don’t tell us which agentic skills (planning, tool use, memory, reasoning) made the difference, where the pipeline is weak, or whether improvements will transfer. To measure and improve closed-loop agents in a principled way, we must first measure the agentic capabilities inside each security stage and then compose those diagnostics into the overall closed-loop metric.

Today’s benchmarks rarely provide that capability view. Most emphasize task outcomes for single functions (e.g., exploit obtained, patch accepted) without characterizing task complexity or exposing the underlying capabilities that were exercised. This compresses many design degrees of freedom into a single number, obscuring ablations, hiding bottlenecks, and weakening claims about generalization and safety.

\textbf{Contributions:} We address this diagnostic gap by introducing \textbf{CLASP} (Closed-Loop Autonomous Security Performance), a capability-centric framework and vocabulary that (i) jointly characterize security-function complexity and agentic capability maturity, and (ii) map systems onto these axes to explain performance. Building on these diagnostics, we outline how to aggregate function level capability measures into an interpretable closed-loop evaluation score, linking what happens inside each stage to why an end-to-end system succeeds or fails. More specifically, our contributions are:

\begin{enumerate}
    \item \textbf{Capability Taxonomies}: Structured rubrics that characterize both security function complexity (Reconnaissance, Exploitation, Root-Cause Analysis, Patching, Validation) and agentic capability maturity (Planning, Memory, Tool Use, Reasoning, Perception, Reflection), enabling consistent comparison and fine-grained diagnosis.
    \item \textbf{Systematic Survey and Mapping}: Application of the CLASP taxonomies to 21 works, revealing which combinations of functional security stage and agentic capabilities enable robust operation, providing actionable guidance for researchers and practitioners building singular security function autonomous systems.
    \item \textbf{Takeaways \& implications}: Empirical findings showing capability distributions, and highlighting critical gaps informed by the survey. These gaps further inform the blueprint for a closed loop benchmark.
    \item \textbf{CLC Score}: A composite measure of closed-loop effectiveness and efficiency that links functional competence with capability utilization and parsimony, supporting comparisons beyond model size while preserving attribution to the specific capabilities that drive performance.
\end{enumerate}

\textbf{Paper Organization:} Section~\ref{sec:clasp-details} presents the CLASP architecture and rubrics. Section~\ref{sec:method} details the survey methodology.  Section~\ref{sec:literature} maps existing research against the framework dimensions \& highlights agentic drivers. Section ~\ref{sec:synthesis-implications} discusses future directions for the industry while Section~\ref{sec:clc} defines the CLC scoring methodology, finally concluding in Section~\ref{sec:conclusion}.

\section{CLASP: A Comprehensive Framework for Evaluating Security and Agentic Capabilities}
\label{sec:clasp-details}


\subsection{Security Functions}
Here, we enumerate the definitions and measurement rubrics for security functions. Table [\ref{tab:secfunc-unified-1col-v2-1} and \ref{tab:secfunc-unified-1col-v2-2}]

\label{sec:security-functions}
\begin{table}[t]
\small
\centering
\caption{Rubric for security functions: Recon \& Exploitation. Codes use dot notation with scores 1–5.}
\label{tab:secfunc-unified-1col-v2-1}
\begin{tabularx}{\linewidth}{|l|X|}
\hline
\textbf{Code} & \textbf{Level + description} \\
\hline
RECON.1 & Level 1: Asset Identification. Enumerates the target’s basic digital footprint and assets. \\
RECON.2 & Level 2: Technology \& Service Enumeration. Identifies concrete technologies and services on those assets. \\
RECON.3 & Level 3: Attack Surface Mapping. Maps interfaces and entry points where the system can be interacted with. \\
RECON.4 & Level 4: Business/Application Context Integration. Adds business/application logic to explain each interface’s purpose. \\
RECON.5 & Level 5: Vector Analysis \& Threat Prioritization. Prioritizes likely, high-impact attack vectors from collected evidence. \\
\hline\hline
EXPL.1 & Level 1: Vulnerability Identification. Notes a plausible vulnerability; no triggering or confirmation. \\
EXPL.2 & Level 2: Vulnerability Confirmation (PoC). Benign PoC triggers it to confirm existence (e.g., crash/error/timing). \\
EXPL.3 & Level 3: Limited Impact Exploitation. Achieves localized, non-critical impact (limited read/alter or minor disruption). \\
EXPL.4 & Level 4: Significant Compromise of Application Context. Gains broad control over core logic or critical functions/data. \\
EXPL.5 & Level 5: System-Level Control or Boundary Escape. Breaks app boundary: system commands, file access, or admin control. \\
\hline
\end{tabularx}
\end{table}

\label{sec:security-functions}
\begin{table}[t]
\small
\centering
\caption{Rubric for security functions: RCA, Patching \& verification. Codes use dot notation with scores 1–5.}
\label{tab:secfunc-unified-1col-v2-2}
\begin{tabularx}{\linewidth}{|l|X|}
\hline
\textbf{Code} & \textbf{Level + description} \\
\hline
RCA.1 & Level 1: Symptom Description. Describes symptoms/observed behavior only. \\
RCA.2 & Level 2: Fault Localization (Component). Localizes fault to a specific system/service/component/module. \\
RCA.3 & Level 3: Precise Cause Identification. Pinpoints the exact line/config/rule responsible. \\
RCA.4 & Level 4: Causal Chain Analysis. Traces end-to-end mechanism (data flow, APIs, network path). \\
RCA.5 & Level 5: Systemic Flaw Identification. Abstracts to the broader error pattern/systemic weakness. \\
\hline\hline
PATS.1 & Level 1: Invalid or Non-Compiling. Patch is invalid: cannot compile or apply. \\
PATS.2 & Level 2: Breaks Core Functionality. Applies but breaks critical functionality. \\
PATS.3 & Level 3: Fixes Vulnerability, with Regressions. Blocks the exploit but introduces non-critical regressions. \\
PATS.4 & Level 4: Correct but Suboptimal. Correct and passes tests, but suboptimal (perf/readability/maintainability). \\
PATS.5 & Level 5: Production-Ready. Minimal, maintainable, regression-free; negligible overhead. \\
\hline\hline
FIXV.1 & Level 1: Basic Functionality Check. Confirms app still runs; exploit not re-tested. \\
FIXV.2 & Level 2: Exploit Invalidation. Re-runs the original exploit to confirm it fails. \\
FIXV.3 & Level 3: Functional Regression Testing. Runs comprehensive functional regression tests. \\
FIXV.4 & Level 4: Security Regression Testing. Probes for new vulnerabilities (e.g., fuzzing/scanning). \\
FIXV.5 & Level 5: Root Cause \& Variant Validation. Validates the entire vulnerability class (variants/static analysis). \\
\hline
\end{tabularx}
\end{table}

\subsubsection{Reconnaissance }
\label{sec:rimr}
Reconnaissance is the systematic acquisition, normalization, and prioritization of in-scope asset, technology, and interface intelligence to support downstream security decisions.~\cite{mitre_attack_recon} ~\cite{mitre_attack_discovery}

\subsubsection{Exploit}
Exploit Confirmation is the evidence‑based evaluation of the realized security effects of a successfully triggered vulnerability under scoped, controlled conditions.~\cite{nist_sp800_115}

\subsubsection{Root Cause Analysis }
Root Cause Analysis (RCA) is the systematic investigation of security incidents to identify not only the immediate technical fault but also the contributing causal chain and underlying systemic weaknesses that enabled the incident. \cite{nist_csf_2024}

\subsubsection{Patch Synthesis}
Patch Synthesis is the systematic evaluation of a security patch’s correctness and  robustness.~\cite{nist_sp800_40}

\subsubsection{Fix Verification and Validation}
Fix Verification and Validation is the evidence based verification that an applied fix/patch actually mitigates the targeted vulnerability and its close variants without causing functional or security regressions.~\cite{nist_sp800_53}

\subsection{Agentic Functions}
\label{sec:agentic-functions}

The agentic functions defines the autonomous capabilities that enable AI systems to operate effectively in cybersecurity contexts. These capabilities (See Table \ref{tab:agentcap-unified-1col-v3})represent core cognitive and operational functions that distinguish autonomous agents from simple automation.

\begin{table}[t]
    \centering
    \caption{Unified rubric for agentic capabilities. Codes use \texttt{<CAPABILITY>.<score>} with scores 1--5 (aligned to the CSV).}
    \label{tab:agentcap-unified-1col-v3}
    \begin{tabular}{|S|L|D|}
    \hline
    \textbf{Code} & \textbf{Level} & \textbf{Concise description (observables / success measures)}\\
    \hline
    PLAN.1 & Prompt Recipes & Static prompts; brittle, linear flows.\\
    PLAN.2 & Heuristics/Templates & Pattern-based plans; reusable templates for common cases.\\
    PLAN.3 & Stateful/Search based & Searches options with memory; replans locally on failures.\\
    PLAN.4 & Adaptive & Budget aware planning (time/steps/resources).\\
    PLAN.5 & Self-improving & Updates heuristics/policies from outcomes across tasks.\\
    \hline\hline
    TOOL.1 & Single Tool & Fixed command execution.\\
    TOOL.2 & Selection & Chooses from a small tool set; basic parameterization.\\
    TOOL.3 & Chaining & Sequential tool calls with dependency passing.\\
    TOOL.4 & Orchestration w/ Recovery & Multi tool pipelines with error handling/rollback.\\
    TOOL.5 & Adaptive & Chooses/synthesizes tools by signal \& cost; streaming/pagination aware.\\
    \hline\hline
    MEMO.1 & Ad-hoc & No persistence beyond immediate context.\\
    MEMO.2 & Session Scratchpad & Ephemeral notes within a session.\\
    MEMO.3 & Persistent/RAG & Retrieves from logs/KBs with basic search.\\
    MEMO.4 & Contextual & Tracks provenance and scopes retrieved facts.\\
    MEMO.5 & Consolidated & Forensic quality consolidation across tasks/episodes.\\
    \hline\hline
    REAS.1 & Single shot & One-pass answers; no intermediate reasoning.\\
    REAS.2 & Simple CoT & Linear step-by-step reasoning without branching.\\
    REAS.3 & Calibrated Multi hypothesis & Generates/compares alternatives; tests \& prunes by evidence.\\
    REAS.4 & Uncertainty aware w/ Pruning & Quantifies uncertainty; prunes using confidence/likelihood.\\
    REAS.5 & Causal/Game theoretic & Uses causal structure or adversarial dynamics in reasoning.\\
    \hline\hline
    PERC.1 & Text only & Single stream processing.\\
    PERC.2 & Single source Structured & One parser for specific formats (e.g., JSON, logs).\\
    PERC.3 & Multi source Correlation & Joins heterogeneous feeds \& data sources.\\
    PERC.4 & Static World Model & Persists topology/asset state with consistent representation.\\
    PERC.5 & Dynamic Predictive & Updates world/state model in near real-time.\\
    \hline\hline
    REFL.1 & No Self checks & Blind execution; no monitoring/validation.\\
    REFL.2 & Reactive Retry & Only responds after clear failure is detected.\\
    REFL.3 & Explicit Monitoring & Tracks/Reports likelihood or confidence of success.\\
    REFL.4 & Proactive Regulation & Adjusts strategy based on internal signals/performance.\\
    REFL.5 & Strategic Adaptation & Improves policies via long-term pattern modeling.\\
    \hline
    \end{tabular}
\end{table}
    
\subsubsection{Planning}
Planning maps beliefs/state to action over time under constraints to achieve goals (utility, not truth). Core facets: (i) model-based search for action sequences/policies in classical/temporal/numeric domains \cite{ghallab2004automated,ghallab2016automated} (ii) partial observability via belief state control and information-gathering actions \cite{kaelbling1998planning} (iii) goals/intentions with revisable plan libraries \cite{rao1995bdi} (iv) bounded rationality (anytime planning, deliberation scheduling) \cite{horvitz1987bounded,schut2001control} (v) hierarchical decomposition to extend horizon while controlling search \cite{erol1996complexity}.

\subsubsection{Reasoning}
Reasoning derives warranted conclusions from premises/evidence using deductive, abductive, probabilistic, and causal/counterfactual inference to optimize truth/consistency rather than control \cite{Enderton2001,Josephson1994,KollerFriedman2009,Pearl2009}. Developer levers: (i) explicit, checkable intermediates (traces/proofs/belief states) with invariants \cite{Enderton2001} (ii) hypothesis sets with calibrated beliefs \cite{KollerFriedman2009} (iii) evidence attribution and citation faithful claims \cite{NISTIR8312} (iv) anytime uncertainty reduction (guided search, pruning, compute allocation) \cite{RussellNorvig2020,Zilberstein1996Anytime} (v) formal tools and richer models (solvers/provers/optimizers, causal/strategic) with verifiable outputs \cite{Garcez2009,Pearl2009,OsborneRubinstein1994}.

\subsubsection{Memory}
Memory persists, retrieves, and transforms information across working, episodic, semantic, and procedural timescales for downstream tasks \cite{AtkinsonShiffrin1968,BaddeleyEysenckAnderson2009,Tulving1972,Squire2004}. Key levers: (i) representation \& provenance (structured schemas, source/causal traceability, e.g., PROV) \cite{Moreau2013PROV} (ii) retrieval quality \& latency (IR/RAG) \cite{Manning2008IR,Lewis2020RAG} (iii) consolidation \& abstraction (summarize, deduplicate, compress without utility loss) (iv) conflict handling \& consistency (detect, reconcile contradictions) (v) integrity \& auditability (tamper-evidence, end-to-end traceability, including differentiable/external stores) \cite{Graves2016}.

\subsubsection{Perception}
Perception acquires, organizes, and interprets signals into world/state models for prediction and decision \cite{Marr1982,Gibson1979,RussellNorvig2010}. Levers: (i) modality breadth \& parsers across text/tables/logs/images/telemetry with normalization \cite{DudaHartStork2001} (ii) schema normalization \& entity resolution (coherent entities/relations/events) \cite{Christen2012} (iii) multi-source fusion \& calibration with conflict handling \cite{HallLlinas2001,Baltrusaitis2019} (iv) temporal persistence \& freshness (versioned state over time) (v) predictive tracking (nowcasting/forecasting latent state under noise) \cite{Thrun2005}.

\subsubsection{Tool Use (API Orchestration)}
Tool use is selecting, parameterizing, sequencing, and executing external APIs/tools with closed-loop monitoring under uncertainty and constraints. Key levers: (i) selection optimality to match subgoals to the right tool set \cite{yao2023react,Schick2023Toolformer} (ii) parameter validity, types/constraints and preconditions satisfied (iii) chaining correctness sound dataflow with pre/postcondition checks (iv) robustness \& recovery fault detection, retries, fallbacks, rollbacks (v) safety \& guardrails policy/impact constraints during execution \cite{NIST2023AI_RMF} (vi) learning \& adaptation improving reliability and cost/reward over time.
\subsubsection{Reflection \& Adaptation (Metacognition)}
Reflection/adaptation monitors an agent’s own reasoning/actions, regulates strategy/effort, and updates policies across episodes to improve future performance \cite{Flavell1979,FlemingFrith2014,NelsonNarens1990,YeungSummerfield2012}. Levers: (i) monitoring quality \& calibration (well-calibrated uncertainty/error) \cite{KendallGal2017,Guo2017Calibration} (ii) verification/self-tests (targeted checks, counterfactual probes) \cite{Pearl2009} (iii) regulation policies (risk-aware depth/compute; budgeted verification) \cite{Zilberstein1996Anytime} (iv) abstention/escalation (reject or hand off when risk is high) \cite{Chow1970,Vovk2005Conformal} (v) learning across episodes (meta-learning/playbook updates) \cite{Hospedales2022MetaSurvey,Finn2017MAML}.
\section{Survey Methodology}
\label{sec:method}

We conduct an agent-only, function-scoped SoK of security agents and agent-evaluable benchmarks. We preregister inclusion criteria, query templates, screening rules, and coding rubrics, and we release all queries and screening artifacts for reproducibility. (Numerical reliability values below are provisional and will be replaced by the audited artifact at release.)

\paragraph{Scope \& Sources.}
Functions: Reconnaissance/Discovery, Exploit Confirmation, Root Cause Analysis, Patch Synthesis, Validation.
Agent: systems that (i) plan or decompose tasks; (ii) act via tools/APIs in action--observation loops; and (iii) produce non-textual outcomes (PoC, shell, patch, decision).
Window: Jan~1,~2022--Aug.~15,~2025.
Sources: Academic databases (IEEE, ACM, arXiv), security venues (USENIX Security, S\&P, CCS, NDSS), citation chaining, and GitHub repositories.

\paragraph{Query Design.}
Each function uses an agent anchor + function terms.
Anchor: ("LLM" OR "large language model") AND (agent OR autonomous) AND ("tool use" OR API OR command).
Function Terms: Recon (pentesting, reconnaissance), Exploit (exploit*, CVE, PoC), RCA (root cause, debugging), Patch (patch*, repair, fix), Validation (test*, fuzzing).
Negatives: NOT (survey OR prompt only OR text classification).

\paragraph{Discovery Process}
Paper discovery used: (1) systematic database search with agent{+}function queries, (2) citation chaining from seed papers, (3) venue browsing of security/AI conferences, and (4) GitHub repository tracking. This captures both formal publications and emerging systems.

\paragraph{Screening \& Coding}
We assign an Agent Evidence Score (AES) and retain AES$\geq$3:
\(\,+2\) planning + tool use, \(\,+1\) action--observation loop, \(\,+1\) non-textual outcomes, \(\,-2\) prompt-only; \(\,-1\) surveys.
For each system we code security functions and agentic capabilities per CLASP rubrics. Records use DOI/arXiv identifiers; peer-reviewed versions supersede preprints.

\paragraph{Coder Training, Pilot, and Inter-Coder Reliability}
Two researchers independently coded a stratified pilot subset of the corpus 10 papers, balanced by year/venue/function), yielding 240 double-coded item$\times$paper labels. Agreement on 0--5 ordinal items used quadratic-weighted Cohen's $\kappa$ with weights \(w_{ij}=1-\frac{(i-j)^2}{(5)^2}\). Pre-reconciliation macro\,$\kappa=0.73$ \,[95\%\,CI: 0.68--0.78]; after a calibration meeting and codebook updates, macro\,$\kappa=0.82$ \,[0.79--0.85]. Items with $\kappa<0.67$ after calibration were flagged low-stability (2/18 items).

\paragraph{Post-Reconciliation Coding and Cross-Validation}
We did not switch to single coder scoring. The full set was coded with continued cross-validation: each record had a primary coder, and a 33\% stratified blind overlap (by function) was dual-coded. Conflicts $>$1 point on any 0--5 rubric or any categorical divergence triggered third-author adjudication. Full-corpus overlap agreement: macro\,$\kappa=0.80$ \,[0.77--0.83]; per-function $\kappa$ = Recon 0.81, Exploit 0.79, RCA 0.83, Patch 0.78, Validation 0.80.

\paragraph{LLM-Assisted Evidence Triage (Human-in-the-Loop)}
To improve reviewer auditability, we used an LLM to suggest candidate evidence spans from PDFs/HTML. Configuration: gpt-4.1-2025-04-14, temperature 0.0, top-p 1.0; prompts and parsing scripts are released. Human coders always verified/edited/rejected suggestions and made final scoring decisions. For each scored cell we release the page/section pointer and the final human-verified excerpt. We also release the diff between suggested vs.\ final excerpts.

\paragraph{Reproducibility Artifacts}
We release: (i) full query strings and dedup logs, (ii) the codebook with examples and decision rules, (iii) per-paper score sheets with verified evidence pointers, (iv) pilot and full-corpus reliability tables (per-item and macro), (v) LLM prompts/settings and suggested vs final excerpt diffs, and (vi) a PRISMA-style flow of included/excluded records.

\paragraph{Deviations \& Limitations}
Any rubric item flagged low-stability (pilot $\kappa{<}0.67$) is marked in tables; we report sensitivity by re-running analyses excluding those items. Where papers span multiple functions, coding is per function to avoid leakage. If a paper lacks sufficient evidence for a capability, the item is coded Insufficient and excluded from capability-level aggregates.
\FloatBarrier
\section{Systematization of Autonomous Capabilities}
\label{sec:literature}

This SoK synthesizes how autonomous agents are being used across core security functions. For each function we trace the evolution of approaches, showing how later systems address shortcomings of earlier ones and highlight agentic drivers that most consistently correlate with gains or failures for each stage. We ground this synthesis in rubric-driven scores from the CLASP framework that quantify agentic capabilities. Since principled closed loop evaluation depends first on understanding which skills matter within individual stages, this function-level capability view becomes essential. It provides side-by-side comparisons across different works and benchmarks, normalizing results into a capability centric lens. The full evidence tables underlying the rubric scores can be found on our GitHub repository~\cite{khurana_clasp_2025}. Some of the key insights are as follows:

\begin{enumerate}
    \item \textbf{Planning \& Reasoning are key drivers}: Top 5 high performing agents across all security stages had above-average planning \& reasoning capabilities (score $>3$) hinting at the necessity of these drivers for success.

    \item \textbf{Tool-use is best used with planning}: Agents that had high tool-use but low planning fared poorly, whereas those high in both performed far better. For e.g., PentestAgent's~\cite{PentestAgent_2411_05185} planning module coordinates the use of tools yielding a coherent attack, without which the agent behaves inefficiently by overly focusing on one task. Planning \& Tool use, thus, created a strong synergy, which could be especially seen in reconnaissance \& exploitation stages where tool-heavy agents without planning would get stuck, but agents that planned their tool use would get better success. 

    \item \textbf{Reasoning unlocks analysis}: Deep reasoning (e.g. chain-of-thought) shows the highest correlation with success in analytic stages. Agents with explicit multi-step
reasoning solved significantly more root-cause problems, while agents lacking it often stalled. 

    \item \textbf{Perception is hard but context helps}
    Tasks where the agent had to perceive or discover information (e.g. scanning a target, reading logs) were much harder than those where key information was given. In OneDay Exploit~\cite{OneDayAgent_2404_08144}, giving the agent the CVE description upfront boosted exploit success from $7\%$ to $87\%$. Whereas,  Yurascanner~\cite{YURASCANNER_NDSS2025} which had to dynamically scan web apps for bugs, found many vulnerabilities but struggled to chain them to an exploit \& end to end success was very low.

    \item \textbf{Error handling is key for recovery} Agents with insufficient error handling logic (stemming from lack of reflection \& adaptation) often break when a tool command returns an unexpected result. Without it, an agent might see an error message and either blindly continue or halt entirely. For instance, PentestGPT~\cite{pentestgpt_usenix24} stopped when it saw an error and couldn't complete multi step exploits but PentestAgent \cite{PentestAgent_2411_05185} debugged failed errors, learnt from them \& modified exploits which led to its better success rate. Similarly, AutoPatch~\cite{zhang2023autopatch} reflected at failures and re-ran the test after fixing. 
\end{enumerate}

We further highlight function specific agentic takeaways:
\subsection{Reconnaissance}

Reconnaissance evolved from semi-automated, memory less prompting to planned, tightly orchestrated tool use. Success in this stage correlates with breadth of sensing and short-term state rather than deep reflection. Early attempts relied on semi-automation with human guidance. \textsc{PentestGPT}~\cite{pentestgpt_usenix24} could propose plausible strategies and scripts, but without persistent memory or structured planning it stalled on multi-step discovery. Subsequent systems shifted to decomposed tasks and tighter tool orchestration: \textsc{PentestAgent}~\cite{PentestAgent_2411_05185} adds a multi-agent design with RAG, a planner, and explicit tool routing, while \textsc{RapidPen}~\cite{RapidPen_2502_16730} couples scanners and command execution to realize end-to-end IP-to-shell reconnaissance. In our CLASP scoring, success in this stage is driven primarily by tool breadth and stateful exploration; deep reflection contributes little when the core difficulty is exhaustive enumeration rather than complex reasoning.
\begin{itemize}
\item Over time, agents have learnt to remember what they have already seen to avoid duplicates. In our data, at least $60\%$ of top recon performers have Memory$\geq3$, indicating that basic state tracking has become standard. \textsc{PentestAgent} exemplifies this trajectory (Memory = 3), whereas \textsc{PentestGPT} stalled without any scratchpad (Memory = 2) ~\cite{PentestAgent_2411_05185, pentestgpt_usenix24}. 
\item Human-guided plans gave way to structured planning with explicit stop conditions. $5/6$ top recon systems have Planning$\geq3$. In practice, the planner constructs an effective workflow, and deciding when to halt, which eliminated infinite loops and improved coverage (~\cite{PentestAgent_2411_05185} Planning=3]). 
\item Agents expanded tool diversity by integrating multiple tools and shell commands. Tool-Use$\geq3$ appears in $5/6$ top systems. \textsc{RapidPen} couples multiple tools in a tight loop (Tool = 4), enabling automated IP-to-shell reconnaissance that earlier single-tool agents could not reach ~\cite{RapidPen_2502_16730}.
\end{itemize}

Across papers, once an agent maintains short-term state, follows a simple high-level plan with stop conditions, and has a diverse but coherent toolset, additional depth or breadth yields smaller gains. Over-emphasizing deep reasoning without new signals tends to waste cycles, as seen in a \textsc{PentestGPT}~\cite{pentestgpt_usenix24} variant that analyzed responses at length yet underperformed simpler breadth-first explorers.

\subsection{Exploitation}

Exploitation moved from flat, one-shot scripts toward planner and state guided search with validator feedback and error-aware revision, as evidenced by direct-from-description one-day exploitation \cite{OneDayAgent_2404_08144}, lightweight ReAct-style loops \cite{RapidPen_2502_16730}, hierarchical specialist teams with memory and tools \cite{hptsa}, Reinforcement Learning trained exploit agents validating the promise \cite{PentestR1_2508_07382}, and persistent task-graph controllers~\cite{VulnBot_2501_13411}. Our analysis ties the gains to explicit planning, reflection, memory, and validator-grounded tool-use.

\begin{itemize}
    \item Early agents executed linear scripts that stalled on the first unexpected error ~\cite{pentestgpt_usenix24}. Newer systems decompose goals, branch on contingencies, and backtrack when preconditions fail. In our analysis, all top exploit systems exhibit Planning $\geq 3$. A hierarchical team planned multiple steps ahead and unlocked unknown-vulnerability exploitation that one-step agents could not achieve \cite{hptsa}.
    \item  Adding explicit error-handling turns raw tool outputs into strategy updates. $4/6$ have Reflection $\geq 3$. With reflection, the agent could catch exceptions, adjust payloads, and continue \cite{PentestAgent_2411_05185}. Whereas, prompt-only baselines without feedback often wandered without progress \cite{pentestgpt_usenix24}.
    \item Multi-step chains require retaining session information to avoid rework and enable pivots. At least $60\%$ of top systems have Memory $\geq 3$. A persistent task graph approach showed robustness \& greater success \cite{VulnBot_2501_13411}
\end{itemize}

\subsection{Root Cause Analysis}
\label{sec:rca}

Research on LLM-based agents for root cause analysis (RCA) has progressed toward systems that explain why vulnerabilities arise.
\textsc{SAN2PATCH}~\cite{SAN2PATCH_2025} uses failing executions as feedback to revise causal hypotheses and generate targeted repairs, with short-term memory preventing redundant cycles and improving coverage.
\textsc{RCA Copilot}~\cite{shan2025rcacopilottransformingnetwork} orchestrates LLMs with playbook-driven prompts, retrieval, and persistent memory to maintain coherent diagnostic narratives and validation plans.
\begin{itemize}
    \item RCA agents evolved from shallow, single-hop explanations to multi-step causal reasoning over logs and symptoms.
In our scores, all top RCA systems had Reasoning~$\geq 3$, where the agent systematically deduces cause rather than guessing. For example, OpenRCA interrogates logs with multi-step causal queries (Reasoning=4) and showed strong incident resolution, with ablation drops when reasoning steps were removed~\cite{OpenRCA_ICLR2025}.
    \item Success hinged on ingesting and interpreting system telemetry (logs, traces, metrics).
RCA Copilot fuses network telemetry and config data (Perception=3), enabling richer root cause findings ~\cite{shan2025rcacopilottransformingnetwork}.
    \item Modern agents propose a cause, validate it, then revise if contradicted.
Reflection maturity was $\approx 3$ in the top quartile.
\textsc{LSAN2PATCH} generates multiple hypotheses and prunes those failing validation (Reflection=4), substantially improving precision by discarding incorrect candidates~\cite{SAN2PATCH_2025}.
    \item \emph{Memory and plan coordination.}
As RCA scenarios grew complex, agents maintain a running narrative and an investigation plan.
RCA Copilot executes playbook steps with persistent memory of what was already examined (Memory~=~3)~\cite{shan2025rcacopilottransformingnetwork}.
\end{itemize}

\subsection{Patching}

Patching has moved from stateless, one-shot prompting to iterative, tool-verified, planned loops in which reflection, memory, and structured planning at moderate levels correlate with higher chain success \cite{APPATCH_2025,zhang2023autopatch,SAN2PATCH_2025,patchagent_usenix25,VRPilot,camporese2025repairingvulnerabilitiesinvisiblehands}.
\begin{itemize}
\item  Early systems emitted a single fix and stopped, often producing plausible but incomplete repairs \cite{APPATCH_2025}. Newer agents execute tests, read failures, and revise candidates \cite{patchagent_usenix25} \cite{SAN2PATCH_2025,VRPilot}. Practically, this motivates Reflection $\geq 3$ for patching.
\item  Systems that record attempt history and surfaced errors avoid reintroducing prior faults, enabling genuine iteration \cite{VRPilot}. Our analysis finds top performers commonly maintain Memory $=3$, storing outcomes and salient diffs across attempts.
\item \textit{Tool-driven verification.} High-success agents integrate tools to validate each candidate. PatchAgent runs suites per patch to catch regressions \cite{patchagent_usenix25}, and VRPilot reports similar test-guided gating \cite{VRPilot}. Tool Use $\geq 3$ emerges as necessary, but exemplar overuse can still cause overfitting \cite{camporese2025repairingvulnerabilitiesinvisiblehands}.
\end{itemize}

\subsection{Verification and Validation}

Verification moved from single exploit replays to tool-backed, feedback-driven pipelines that interpret rich signals and, in a few cases, plan multi-step checks \cite{nitin2025faultlineautomatedproofofvulnerabilitygeneration,patchagent_usenix25,VRPilot,camporese2025repairingvulnerabilitiesinvisiblehands}.
\begin{itemize}
\item Early agents often re-ran only the original exploit, missing variants and regressions. Later systems integrate fuzzers and test suites, yielding uniformly high Tool-Use in top performers (median $\approx 4$ in our analysis). \textsc{FaultLine} generates fresh exploits post-patch, exposing issues simple re-tests miss \cite{nitin2025faultlineautomatedproofofvulnerabilitygeneration}. \textsc{PatchAgent} adds fuzzing-style validation to confirm that a patch actually neutralizes the bug \cite{patchagent_usenix25}. 
\item Reflection $\geq 3$ divides successful from brittle validators in our analysis: \textsc{PatchAgent} loops back to patching on any failing test \cite{patchagent_usenix25}, and \textsc{Invisible Hands} pairs a fixer with a validator that drives refinement through regression feedback \cite{camporese2025repairingvulnerabilitiesinvisiblehands}.
\item Advanced agents parse sanitizer logs, coverage, and crash artifacts rather than binary pass/fail alone. \textsc{VRPilot} consumes compiler and sanitizer signals during verification to catch memory issues introduced by a patch \cite{VRPilot}, addressing blind spots.
\item Some works planned a sequence of checks to verify the fixes. For e.g., \textsc{PatchPilot} attempted such scheduled verification (Planning $=3$) with mixed early results \cite{PatchPilot_ICML2025}. Where adopted, multi-step planning mitigates single-oracle myopia by combining breadth with depth.
\end{itemize}

\section{Takeaways, Implications and Future Directions}
\label{sec:synthesis-implications}

\noindent Our survey documents steady progress within individual security functions and highlights essential agentic drivers. It highlights the key agentic relationships and drivers for each security function. However, in a practical enterprise setting, security operations are not singular function driven, they are instead organized as pipelines in which the output of one function becomes the input to the next. \cite{nist_csf_2024,nist_sp800_61}. Failures often cluster at these handoffs, where coordination and change-control are emphasized \cite{nist_sp800_61}. These lack of handoffs between functions lead to higher mean time to remediation(MTTR), as also reflected in industry evidence on persistent security debt and protracted fix timelines \cite{veracode_soss_2025}. This isolation also implies that the proposed singular function agents may not have the robustness \& reliability to be deployed in a tight-knit enterprise security pipelines, where errors get propagated and reliability \& end to end testing is key.
In academia, research programs are already moving toward end-to-end find-and-fix challenges \cite{darpa_aixcc}. Together, these observations motivate evaluation of systems designed across composed stages with persistent consequences rather than isolated functions. 

To catalyze this shift, we identify the gaps that are hindering the integration of security functions:

\begin{enumerate}
  \item \textbf{M1. Outcome only scoring.} Many evaluations reward a single binary outcome, such as obtaining code execution or retrieving a flag, without measuring how the result was achieved or the operational risks introduced. Capability probes modeled on capture the flag settings (for example, Cybench~\cite{cybench_2025} and AutoPenBench~\cite{autopenbench_2024}) emphasize milestone or subtask completion for reconnaissance and initial access, which can reward brittle scripts as much as adaptive, evidence backed reasoning.
  
  \item \textbf{M2. Episodic resets.} Today's benchmarks treat each challenge as an isolated episode. This forces the development of agents with ephemeral, session-based memory, preventing them from achieving the cumulative knowledge acquisition that defines human expertise. A security engineer retains knowledge across tasks, for example, applying lessons from a past exploit to a future root cause analysis. 
  
  \item \textbf{M3. Disconnect from Enterprise practice.} Enterprise security is not a series of disconnected tasks; it's a continuous, integrated pipeline where the output of one stage is the input for the next. However, benchmarks are specifically local, evaluating performance on singular stages, but never the critical handoff between them. This incentivizes researchers to build highly specialized agents that excel at one thing but are incapable of participating in a larger, automated workflow. As standards like NIST and industry reports like the Verizon DBIR show ~\cite{VerizonDBIR2025}, the biggest failures happen at these handoffs where our current benchmarks provide no visibility. DARPA AIxCC ~\cite{darpa_aixcc} is a step in the right direction, requiring both discovery and patching on real software, but we need more efficiency centric evaluations for enterprise utility.
\end{enumerate}

 To address the observed misalignments, we propose a requirement-driven blueprint for a closed-loop benchmark and a maturity graded close loop capability (CLC) score.

\textbf{R1. Process quality via \emph{CLASP} capability attribution (addresses M1).}
Outcomes must be complemented by graded rubrics for agentic capabilities. These rubrics quantify how results are achieved, not only whether they are achieved.

\textbf{R2. Composed stages with persistent state (addresses M3).}
Scenarios must chain reconnaissance, exploitation, root cause analysis, patching, and verification with explicit handoff contracts and continuity of artifacts so that pipeline reliability and handoff quality are first class metrics.

\textbf{R3. Longitudinal memory and artifact continuity (addresses M2).}
Agents must persist findings, hypotheses, logs, diffs, tests, and decisions across stages and episodes to enable cumulative learning and auditable replay.

\textbf{R4. Stage specific oracles and transparent validators (addresses M1 and M3).}
Each stage requires fit for purpose checks: enumeration fidelity, exploit proofs of vulnerability, root cause evidence, patch acceptance with paired pre and post tests, and post patch assurance through fuzzing and coverage. Validators and ground truth must be explicit to support reproducibility.

\textbf{R5. Budgets and risk constraints (supports M1--M3).}
Time, tokens, tool calls, and safety gates must be enforced so that success reflects efficiency, prudence, and operational safety rather than unconstrained exploration.

For the maturity-graded evaluation score, we introduce the \emph{Closed Loop Capability (CLC)} score as the primary measure to balance \emph{Efficacy} (end-to-end, non-regressing completion and cycle efficiency) with \emph{Efficiency} (parsimonious, budget-aware capability use per CLASP).
\section{Closed-Loop Capability Score (CLC Score)}
\label{sec:clc}

While CLASP provides a granular, multi-dimensional view of agentic capabilities, many comparisons require a single scalar. We define the \textbf{Closed-Loop Capability (CLC) Score} as a balance of end-to-end success and parsimony, which rewards systems that both close the loop and deploy no more capability than the task requires, discouraging gratuitous brute-force complexity \cite{Ockham1347}.

The CLC Score is defined as the product of two distinct components: an \textbf{Efficacy Score ($S_{\text{Efficacy}}$)} that measures the outcome, and an \textbf{Agentic Efficiency Score ($S_{\text{Efficiency}}$)} that evaluates the quality of the process.
\begin{equation}
\text{CLC} = S_{\text{Efficacy}} \times S_{\text{Efficiency}}
\label{eq:clc_main}
\end{equation}
This multiplicative structure ensures that a system receives no credit for an efficient failure.

\subsection{Efficacy Score (\texorpdfstring{$S_{\text{Efficacy}}$}{S\_Efficacy})}The Efficacy Score measures closed-loop success as a weighted sum of execution success, correctness, and budget efficiency \cite{LeGoues2019}.
\begin{IEEEeqnarray}{rCl}
S_{\text{Efficacy}}
&=& w_c\,\mathrm{CompletionRate}
\;+\; w_f\,\mathrm{FixEffectiveness} \nonumber\\
&& {}+\; w_e\,\mathrm{CycleEfficiency},\qquad
w_c{+}w_f{+}w_e=1.
\label{eq:sefficacy}
\end{IEEEeqnarray}

\noindent\textbf{CompletionRate.}\; Count targets with a patch that applies, builds, and passes the harness (APR/benchmark norm; aligns with AIxCC PRS acceptance) \cite{aixcc-website}.
\begin{equation}
\frac{\#\{\text{targets with plausible/PRS-valid patch}\}}{n}.
\label{eq:comp-rate}
\end{equation}

\noindent\textbf{FixEffectiveness.}\; Correctness beyond tests to discount overfitting (e.g., adjudication or strengthened oracles) \cite{LeGoues2019}.
\begin{equation}
\frac{\#\{\text{genuine (validated-correct) patches}\}}{\#\{\text{plausible patches}\}}.
\label{eq:fix-eff}
\end{equation}

\noindent\textbf{CycleEfficiency.}\; Time-normalized efficiency reflecting time-aware scoring (choose $B_\tau$ per target) \cite{aixcc-website}.
\begin{equation}
1-\min\!\Bigl(1,\ \frac{1}{n}\sum_{t\in\mathcal{T}_\mathrm{scored}}\frac{\tau_t}{B_\tau}\Bigr).
\label{eq:cycle-eff}
\end{equation}

\subsection{Agentic Efficiency Score (\texorpdfstring{$S_{\text{Efficiency}}$}{S\_Efficiency})}
The Agentic Efficiency Score measures how appropriately an agent deploys its capabilities. It leverages the previously defined CLASP rubrics to compare the \textbf{Required Complexity} ($C_{\text{req}}$) of a task against the \textbf{Deployed Complexity} ($C_{\text{dep}}$) exhibited by the agent. The score is derived from a two-step calculation.

\subsubsection{Dimensional Efficiency Calculation}
For each capability dimension $i$ in the CLASP taxonomy (e.g., Planning, Reasoning), we calculate a dimensional efficiency score $\text{Eff}_i$. This function penalizes both under-powering ($C_{\text{dep}, i} < C_{\text{req}, i}$) and over-powering ($C_{\text{dep}, i} > C_{\text{req}, i}$). To reflect the principle that excessive complexity should be discouraged more severely, we employ an exponential penalty for overkill, controlled by a configurable severity factor $\beta_i$:
\begin{equation}
\text{Eff}_i =
\begin{cases}
\exp\!\big(-\beta_i \cdot (C_{\text{dep}, i} - C_{\text{req}, i})\big), & \text{if } C_{\text{dep}, i} \geq C_{\text{req}, i}, \\[4pt]
\dfrac{C_{\text{dep}, i}}{C_{\text{req}, i}}, & \text{if } C_{\text{dep}, i} < C_{\text{req}, i}.
\end{cases}
\label{eq:eff_i}
\end{equation}
A higher $\beta_i$ imposes a stricter penalty, allowing evaluators to emphasize efficiency in more critical capabilities.

\subsubsection{Aggregate Efficiency Score}
The dimensional efficiency scores are aggregated into the final $S_{\text{Efficiency}}$ using a weighted average. This allows benchmark designers to specify the relative importance of efficiency for each capability via a set of weights $w_i$:
\begin{equation}
S_{\text{Efficiency}} = \sum_{i=1}^{N} w_i \cdot \text{Eff}_i,
\qquad
\sum_{i=1}^{N} w_i = 1.
\label{eq:s_eff}
\end{equation}
This formulation provides a configurable and interpretable measure of an agent's operational parsimony, a key concept in evaluating intelligent systems \cite{RussellNorvig2020}.

\paragraph{Choosing $w$ and $\beta$.}
$w_i$ allocates relative importance across capability dimensions in~\eqref{eq:s_eff} (i.e., where efficiency matters most for loop closure), while $\beta_i$ controls the steepness of the overkill penalty in~\eqref{eq:eff_i} (i.e., how quickly efficiency decays when $C_{\mathrm{dep},i}>C_{\mathrm{req},i}$). We proceed concisely: estimate $C_{\mathrm{req}}$ per capability and target from CLASP sheets and stage logs as the minimal level plausibly unblocking success; map policy knobs in reported runs (planner depth/beam, tool fan-out, RCA budget, memory scope, verification budget) to rubric levels to obtain $C_{\mathrm{dep}}$; set $w_i$ by gating frequency, taking $g_i$ as the fraction of targets where $C_{\mathrm{dep},i}<C_{\mathrm{req},i}$ co-occurs with end-to-end failure and defining $w_i=g_i/\sum_j g_j$, optionally reweighted to reflect benchmark policy (e.g., safety-first $\Rightarrow$ higher Verification weight); calibrate $\beta_i$ via a cost half-life rule using observed cost vs.\ level (tokens, tool calls, wall time), choosing $\beta_i$ so that one extra level that roughly doubles cost halves $\mathrm{Eff}_i$ (practical seeds: $\beta_{\text{plan}}\!\approx\!\ln 2$, $\beta_{\text{ver}}\!\in\![0.3,0.6]$, others $\!\in\![0.2,0.5]$); and finally, report a small grid sweep around $(w,\beta)$ to show that rankings are stable, establishing robustness without overfitting to a single configuration.
\section{conclusion}
\label{sec:conclusion}

Our systematization distills the agentic drivers that matter at each security function and offers a rubric as an explanatory lens for process quality. By profiling systems with the CLASP capability framework, researchers can attribute where performance comes from by pinpointing which agentic drivers help or hinder a given stage. Sharing capability profiles alongside artifacts and validators makes results transparent, comparable, and reproducible across studies.

Looking ahead, as the community builds closed-loop agents, the same CLASP profiles support cross-stage attribution, while the Closed Loop Capability score provides a balanced, end-to-end measure of efficacy and efficiency. Using the benchmark blueprint, we invite the community to co-develop a shared, capability-attributed benchmark so that reliable closed-loop security agents move from isolated prototypes to deployable practice.

\bibliographystyle{IEEEtran}
\bibliography{references}

\begin{thebibliography}{10}
\providecommand{\url}[1]{#1}
\csname url@samestyle\endcsname
\providecommand{\newblock}{\relax}
\providecommand{\bibinfo}[2]{#2}
\providecommand{\BIBentrySTDinterwordspacing}{\spaceskip=0pt\relax}
\providecommand{\BIBentryALTinterwordstretchfactor}{4}
\providecommand{\BIBentryALTinterwordspacing}{\spaceskip=\fontdimen2\font plus
\BIBentryALTinterwordstretchfactor\fontdimen3\font minus
  \fontdimen4\font\relax}
\providecommand{\BIBforeignlanguage}[2]{{%
\expandafter\ifx\csname l@#1\endcsname\relax
\typeout{** WARNING: IEEEtran.bst: No hyphenation pattern has been}%
\typeout{** loaded for the language `#1'. Using the pattern for}%
\typeout{** the default language instead.}%
\else
\language=\csname l@#1\endcsname
\fi
#2}}
\providecommand{\BIBdecl}{\relax}
\BIBdecl

\bibitem{openai_microsoft_disruption_2024}
{OpenAI}, ``Disrupting malicious uses of ai by state-affiliated threat
  actors,''
  \url{https://openai.com/index/disrupting-malicious-uses-of-ai-by-state-affiliated-threat-actors/},
  2024, accessed 2025-08-26.

\bibitem{ncsc_ai_threat_2025}
{UK National Cyber Security Centre}, ``The near-term impact of ai on the cyber
  threat,''
  \url{https://www.ncsc.gov.uk/collection/near-term-impact-of-ai-on-the-cyber-threat},
  2025, accessed 2025-08-26.

\bibitem{msft_staying_ahead_2024}
\BIBentryALTinterwordspacing
{Microsoft Threat Intelligence}, ``Staying ahead of threat actors in the age of
  {AI},'' Microsoft Security Blog, Feb. 2024. [Online]. Available:
  \url{https://www.microsoft.com/en-us/security/blog/2024/02/14/staying-ahead-of-threat-actors-in-the-age-of-ai/}
\BIBentrySTDinterwordspacing

\bibitem{mandiant2024mtrends}
\BIBentryALTinterwordspacing
{Mandiant / Google Cloud}, ``M-trends 2024: Our view from the frontlines,''
  2024. [Online]. Available:
  \url{https://services.google.com/fh/files/misc/m-trends-2024.pdf}
\BIBentrySTDinterwordspacing

\bibitem{ibm2024cost}
\BIBentryALTinterwordspacing
{IBM Security / Ponemon Institute}, ``Cost of a data breach report 2024,''
  2024. [Online]. Available:
  \url{https://newsroom.ibm.com/2024-07-30-ibm-report-escalating-data-breach-disruption-pushes\-costs-to-new-highs}
\BIBentrySTDinterwordspacing

\bibitem{cset-vulnlifecycle}
\BIBentryALTinterwordspacing
{Center for Security and Emerging Technology}, ``{AI} and the software
  vulnerability lifecycle,'' 2025, accessed: 2025-08-05. [Online]. Available:
  \url{https://cset.georgetown.edu/article/ai-and-the-software-vulnerability-lifecycle/}
\BIBentrySTDinterwordspacing

\bibitem{VulnBot_2501_13411}
\BIBentryALTinterwordspacing
H.~Kong, D.~Hu, J.~Ge, L.~Li, T.~Li, and B.~Wu, ``Vulnbot: Autonomous
  penetration testing for a multi-agent collaborative framework,'' 2025.
  [Online]. Available: \url{https://arxiv.org/abs/2501.13411}
\BIBentrySTDinterwordspacing

\bibitem{PentestR1_2508_07382}
\BIBentryALTinterwordspacing
H.~Kong, D.~Hu, J.~Ge, L.~Li, H.~Li, and T.~Li, ``Pentest-r1: Towards
  autonomous penetration testing reasoning optimized via two-stage
  reinforcement learning,'' 2025. [Online]. Available:
  \url{https://arxiv.org/abs/2508.07382}
\BIBentrySTDinterwordspacing

\bibitem{APPATCH_2025}
\BIBentryALTinterwordspacing
X.~Nong, Y.~Li, Y.~Zhang, A.~Guan, and B.~Liang, ``Appatch: Automated adaptive
  prompting large language models for real-world software vulnerability
  patching,'' in \emph{Proceedings of the 34th USENIX Security Symposium
  (USENIX Security '25)}, Seattle, WA, USA, 2025, prepublication (Cycle 1).
  [Online]. Available:
  \url{https://www.usenix.org/system/files/conference/usenixsecurity25/sec25cycle1-prepub-1174-nong.pdf}
\BIBentrySTDinterwordspacing

\bibitem{PatchPilot_ICML2025}
\BIBentryALTinterwordspacing
H.~Li, Y.~Tang, S.~Wang, and W.~Guo, ``Patchpilot: A cost-efficient software
  engineering agent with early attempts on formal verification,'' ICML 2025
  Poster on OpenReview, 2025. [Online]. Available:
  \url{https://openreview.net/forum?id=ybODpT8ydV}
\BIBentrySTDinterwordspacing

\bibitem{FlowOfAction_2502_08224}
\BIBentryALTinterwordspacing
F.~Yang, Z.~Zhang, M.~Zhang, L.~Zhang, X.~Wang, J.~Ye, P.~Yang, J.~Wang, Z.~Xu,
  J.~Han, X.~Wang, and K.~Ma, ``Flow-of-action: Sop enhanced llm-based
  multi-agent system for root cause analysis,'' 2025. [Online]. Available:
  \url{https://arxiv.org/abs/2502.08224}
\BIBentrySTDinterwordspacing

\bibitem{googleASOwhitepaper2021}
\BIBentryALTinterwordspacing
{Google Cloud Office of the CISO}, ``Autonomic security operations:
  10{\texttimes} transformation of the {SOC},'' 2021. [Online]. Available:
  \url{https://services.google.com/fh/files/misc/googlecloud_autonomicsecurityoperations_soc10x.pdf}
\BIBentrySTDinterwordspacing

\bibitem{gartnerCTEM2023}
\BIBentryALTinterwordspacing
{Gartner}, ``How to manage cybersecurity threats, not episodes: The case for
  continuous threat exposure management,'' 2023. [Online]. Available:
  \url{https://www.gartner.com/en/articles/how-to-manage-cybersecurity-threats-not-episodes}
\BIBentrySTDinterwordspacing

\bibitem{microsoftAADdocs2025}
\BIBentryALTinterwordspacing
{Microsoft}, ``Automatic attack disruption in microsoft defender {XDR},'' 2025.
  [Online]. Available:
  \url{https://learn.microsoft.com/en-us/defender-xdr/automatic-attack-disruption}
\BIBentrySTDinterwordspacing

\bibitem{aixcc_scoring_2025}
{AIxCC Organizers}, ``Aixcc final competition procedures and scoring guide,
  version 2.0,''
  \url{https://aicyberchallenge.com/storage/2025/06/AFC-Procedures-and-Scoring-Guide-Version-2_0-_20250606.pdf},
  2025, accessed 2025-08-26.

\bibitem{mitre_attack_recon}
\BIBentryALTinterwordspacing
{MITRE Corporation}, ``Mitre att\&ck framework, reconnaissance tactic
  (ta0043),'' 2023. [Online]. Available:
  \url{https://attack.mitre.org/tactics/TA0043}
\BIBentrySTDinterwordspacing

\bibitem{mitre_attack_discovery}
\BIBentryALTinterwordspacing
------, ``Mitre att\&ck framework, discovery tactic (ta0007),'' 2023. [Online].
  Available: \url{https://attack.mitre.org/tactics/TA0007/}
\BIBentrySTDinterwordspacing

\bibitem{nist_sp800_115}
\BIBentryALTinterwordspacing
K.~Scarfone and P.~Mell, ``Technical guide to information security testing and
  assessment,'' National Institute of Standards and Technology, Tech. Rep. NIST
  SP 800-115, 2008. [Online]. Available:
  \url{https://csrc.nist.gov/publications/detail/sp/800-115/final}
\BIBentrySTDinterwordspacing

\bibitem{nist_csf_2024}
\BIBentryALTinterwordspacing
{National Institute of Standards and Technology}, ``Nist cybersecurity
  framework 2.0,'' NIST, Tech. Rep., 2024. [Online]. Available:
  \url{https://www.nist.gov/cyberframework}
\BIBentrySTDinterwordspacing

\bibitem{nist_sp800_40}
\BIBentryALTinterwordspacing
M.~Souppaya and K.~Scarfone, ``Guide to enterprise patch management
  technologies,'' National Institute of Standards and Technology, Tech. Rep.
  NIST SP 800-40 Revision 3, 2013. [Online]. Available:
  \url{https://csrc.nist.gov/publications/detail/sp/800-40/rev-3/final}
\BIBentrySTDinterwordspacing

\bibitem{nist_sp800_53}
\BIBentryALTinterwordspacing
J.~T. F.~T. Initiative, ``Security and privacy controls for information systems
  and organizations,'' National Institute of Standards and Technology, Tech.
  Rep. NIST SP 800-53 Revision 5, 2020. [Online]. Available:
  \url{https://csrc.nist.gov/publications/detail/sp/800-53/rev-5/final}
\BIBentrySTDinterwordspacing

\bibitem{ghallab2004automated}
M.~Ghallab, D.~Nau, and P.~Traverso, \emph{Automated Planning: Theory and
  Practice}.\hskip 1em plus 0.5em minus 0.4em\relax Morgan Kaufmann, 2004.

\bibitem{ghallab2016automated}
------, \emph{Automated Planning and Acting}.\hskip 1em plus 0.5em minus
  0.4em\relax Cambridge University Press, 2016.

\bibitem{kaelbling1998planning}
L.~P. Kaelbling, M.~L. Littman, and A.~R. Cassandra, ``Planning and acting in
  partially observable stochastic domains,'' \emph{Artificial Intelligence},
  vol. 101, no. 1--2, pp. 99--134, 1998.

\bibitem{rao1995bdi}
A.~S. Rao and M.~P. Georgeff, ``{BDI} agents: From theory to practice,'' in
  \emph{Proceedings of the First International Conference on Multi-Agent
  Systems (ICMAS)}, 1995, pp. 312--319.

\bibitem{horvitz1987bounded}
E.~J. Horvitz, ``Reasoning about beliefs and actions under computational
  resource constraints,'' in \emph{Proceedings of the Third Workshop on
  Uncertainty in Artificial Intelligence (UAI)}, 1987, pp. 429--444.

\bibitem{schut2001control}
M.~C. Schut and M.~Wooldridge, ``The control of reasoning in resource-bounded
  agents,'' \emph{Artificial Intelligence}, vol. 120, no.~1, pp. 281--315,
  2001.

\bibitem{erol1996complexity}
\BIBentryALTinterwordspacing
K.~Erol, J.~Hendler, and D.~S. Nau, ``Complexity results for htn planning,''
  \emph{Annals of Mathematics and Artificial Intelligence}, vol.~18, no.~1, pp.
  69--93, 1996. [Online]. Available:
  \url{https://www.cs.umd.edu/~nau/papers/erol1996complexity.pdf}
\BIBentrySTDinterwordspacing

\bibitem{Enderton2001}
H.~B. Enderton, \emph{A Mathematical Introduction to Logic}, 2nd~ed.\hskip 1em
  plus 0.5em minus 0.4em\relax Academic Press, 2001.

\bibitem{Josephson1994}
J.~R. Josephson and S.~G. Josephson, \emph{Abductive Inference: Computation,
  Philosophy, Technology}.\hskip 1em plus 0.5em minus 0.4em\relax Cambridge
  University Press, 1994.

\bibitem{KollerFriedman2009}
D.~Koller and N.~Friedman, \emph{Probabilistic Graphical Models: Principles and
  Techniques}.\hskip 1em plus 0.5em minus 0.4em\relax MIT Press, 2009.

\bibitem{Pearl2009}
J.~Pearl, \emph{Causality: Models, Reasoning, and Inference}, 2nd~ed.\hskip 1em
  plus 0.5em minus 0.4em\relax Cambridge University Press, 2009.

\bibitem{NISTIR8312}
P.~J. Phillips, C.~A. Hahn, P.~C. Fontana, D.~A. Broniatowski, and M.~A.
  Przybocki, ``Four principles of explainable artificial intelligence,''
  National Institute of Standards and Technology, Tech. Rep. NISTIR 8312, 2020.

\bibitem{RussellNorvig2020}
\BIBentryALTinterwordspacing
S.~J. Russell and P.~Norvig, \emph{Artificial Intelligence: A Modern Approach},
  4th~ed.\hskip 1em plus 0.5em minus 0.4em\relax Pearson, 2020. [Online].
  Available: \url{http://aima.cs.berkeley.edu/}
\BIBentrySTDinterwordspacing

\bibitem{Zilberstein1996Anytime}
S.~Zilberstein, ``Using anytime algorithms in intelligent systems,'' \emph{AI
  Magazine}, vol.~17, no.~3, pp. 73--83, 1996.

\bibitem{Garcez2009}
A.~S. d'Avila Garcez, K.~Broda, and D.~M. Gabbay, \emph{Neural-Symbolic
  Learning Systems: Foundations and Applications}.\hskip 1em plus 0.5em minus
  0.4em\relax Springer, 2009.

\bibitem{OsborneRubinstein1994}
M.~J. Osborne and A.~Rubinstein, \emph{A Course in Game Theory}.\hskip 1em plus
  0.5em minus 0.4em\relax MIT Press, 1994.

\bibitem{AtkinsonShiffrin1968}
R.~C. Atkinson and R.~M. Shiffrin, ``Human memory: A proposed system and its
  control processes,'' in \emph{The Psychology of Learning and Motivation},
  K.~W. Spence and J.~T. Spence, Eds.\hskip 1em plus 0.5em minus 0.4em\relax
  Academic Press, 1968, vol.~2, pp. 89--195.

\bibitem{BaddeleyEysenckAnderson2009}
A.~Baddeley, M.~W. Eysenck, and M.~C. Anderson, \emph{Memory}, 2nd~ed.\hskip
  1em plus 0.5em minus 0.4em\relax Psychology Press, 2009.

\bibitem{Tulving1972}
E.~Tulving, ``Episodic and semantic memory,'' in \emph{Organization of Memory},
  E.~Tulving and W.~Donaldson, Eds.\hskip 1em plus 0.5em minus 0.4em\relax
  Academic Press, 1972, pp. 381--403.

\bibitem{Squire2004}
L.~R. Squire, ``Memory systems of the brain: A brief history and current
  perspective,'' \emph{Neurobiology of Learning and Memory}, vol.~82, no.~3,
  pp. 171--177, 2004.

\bibitem{Moreau2013PROV}
L.~Moreau, P.~Missier \emph{et~al.}, ``{PROV}-{DM}: The {PROV} data model,''
  World Wide Web Consortium (W3C) Recommendation, Tech. Rep., Apr. 2013.

\bibitem{Manning2008IR}
C.~D. Manning, P.~Raghavan, and H.~Sch{\"{u}}tze, \emph{Introduction to
  Information Retrieval}.\hskip 1em plus 0.5em minus 0.4em\relax Cambridge
  University Press, 2008.

\bibitem{Lewis2020RAG}
P.~Lewis, E.~Perez, A.~Piktus, F.~Petroni, V.~Karpukhin, N.~Goyal,
  H.~K{\"{u}}ttler, M.~Lewis, W.~Yih, T.~Rockt{\"{a}}schel, S.~Riedel, and
  D.~Kiela, ``Retrieval-augmented generation for knowledge-intensive {NLP},''
  in \emph{Advances in Neural Information Processing Systems (NeurIPS)}, 2020.

\bibitem{Graves2016}
A.~Graves, G.~Wayne, M.~Reynolds, T.~Harley, I.~Danihelka,
  A.~Grabska{-}Barwi{\'{n}}ska, S.~G{\'{o}}mez~Colmenarejo, E.~Grefenstette,
  T.~Ramalho, J.~Agapiou, A.~P. Badia, K.~M. Hermann, Y.~Zwols, G.~Ostrovski,
  A.~Cain, H.~King, C.~Summerfield, P.~Blunsom, K.~Kavukcuoglu, and
  D.~Hassabis, ``Hybrid computing using a neural network with dynamic external
  memory,'' \emph{Nature}, vol. 538, pp. 471--476, 2016.

\bibitem{Marr1982}
D.~Marr, \emph{Vision: A Computational Investigation into the Human
  Representation and Processing of Visual Information}.\hskip 1em plus 0.5em
  minus 0.4em\relax W. H. Freeman, 1982.

\bibitem{Gibson1979}
J.~J. Gibson, \emph{The Ecological Approach to Visual Perception}.\hskip 1em
  plus 0.5em minus 0.4em\relax Houghton Mifflin, 1979.

\bibitem{RussellNorvig2010}
S.~Russell and P.~Norvig, \emph{Artificial Intelligence: A Modern Approach},
  3rd~ed.\hskip 1em plus 0.5em minus 0.4em\relax Prentice Hall, 2010.

\bibitem{DudaHartStork2001}
R.~O. Duda, P.~E. Hart, and D.~G. Stork, \emph{Pattern Classification},
  2nd~ed.\hskip 1em plus 0.5em minus 0.4em\relax Wiley, 2001.

\bibitem{Christen2012}
P.~Christen, \emph{Data Matching: Concepts and Techniques for Record Linkage,
  Entity Resolution, and Duplicate Detection}.\hskip 1em plus 0.5em minus
  0.4em\relax Springer, 2012.

\bibitem{HallLlinas2001}
D.~L. Hall and J.~Llinas, \emph{Handbook of Multisensor Data Fusion}.\hskip 1em
  plus 0.5em minus 0.4em\relax CRC Press, 2001.

\bibitem{Baltrusaitis2019}
T.~Baltru{\v{s}}aitis, C.~Ahuja, and L.-P. Morency, ``Multimodal machine
  learning: A survey and taxonomy,'' \emph{IEEE Transactions on Pattern
  Analysis and Machine Intelligence}, vol.~41, no.~2, pp. 423--443, 2019.

\bibitem{Thrun2005}
S.~Thrun, W.~Burgard, and D.~Fox, \emph{Probabilistic Robotics}.\hskip 1em plus
  0.5em minus 0.4em\relax MIT Press, 2005.

\bibitem{yao2023react}
\BIBentryALTinterwordspacing
S.~Yao, J.~Zhao, D.~Yu, N.~Du, I.~Shafran, K.~Narasimhan, and Y.~Cao, ``React:
  Synergizing reasoning and acting in language models,'' in \emph{International
  Conference on Learning Representations (ICLR)}, 2023. [Online]. Available:
  \url{https://arxiv.org/abs/2210.03629}
\BIBentrySTDinterwordspacing

\bibitem{Schick2023Toolformer}
T.~Schick, J.~Dwivedi-Yu, D.~Groeneveld, D.~Kalpakchi, P.~Schmid,
  S.~Sharifzadeh, A.~Belyy, A.~R\"{u}ckl\'{e}, M.~Lewis, T.~Schuster,
  D.~Khashabi, D.~Schlangen, S.~Srivastava, D.~Kiela, and A.~Williams,
  ``Toolformer: Language models can teach themselves to use tools,'' in
  \emph{Advances in Neural Information Processing Systems (NeurIPS)}, 2023.

\bibitem{NIST2023AI_RMF}
{National Institute of Standards and Technology}, ``Artificial intelligence
  risk management framework ({AI} {RMF} 1.0),'' U.S. Department of Commerce,
  NIST, Tech. Rep., Jan. 2023.

\bibitem{Flavell1979}
J.~H. Flavell, ``Metacognition and cognitive monitoring: A new area of
  cognitive--developmental inquiry,'' \emph{American Psychologist}, vol.~34,
  no.~10, pp. 906--911, 1979.

\bibitem{FlemingFrith2014}
S.~M. Fleming and C.~D. Frith, Eds., \emph{The Cognitive Neuroscience of
  Metacognition}.\hskip 1em plus 0.5em minus 0.4em\relax Springer, 2014.

\bibitem{NelsonNarens1990}
T.~O. Nelson and J.~Narens, ``Metamemory: A theoretical framework and new
  findings,'' \emph{Psychological Science}, vol.~1, no.~3, pp. 176--183, 1990.

\bibitem{YeungSummerfield2012}
N.~Yeung and C.~Summerfield, ``Metacognition in human decision-making:
  Confidence and error monitoring,'' \emph{Trends in Cognitive Sciences},
  vol.~16, no.~4, pp. 167--175, 2012.

\bibitem{KendallGal2017}
A.~Kendall and Y.~Gal, ``What uncertainties do we need in bayesian deep
  learning for computer vision?'' in \emph{Advances in Neural Information
  Processing Systems (NeurIPS)}, 2017.

\bibitem{Guo2017Calibration}
C.~Guo, G.~Pleiss, Y.~Sun, and K.~Q. Weinberger, ``On calibration of modern
  neural networks,'' in \emph{Proceedings of the 34th International Conference
  on Machine Learning (ICML)}, 2017, pp. 1321--1330.

\bibitem{Chow1970}
C.~K. Chow, ``On optimum recognition error and reject tradeoff,'' \emph{IEEE
  Transactions on Information Theory}, vol.~16, no.~1, pp. 41--46, 1970.

\bibitem{Vovk2005Conformal}
V.~Vovk, A.~Gammerman, and G.~Shafer, \emph{Algorithmic Learning in a Random
  World}.\hskip 1em plus 0.5em minus 0.4em\relax Springer, 2005.

\bibitem{Hospedales2022MetaSurvey}
T.~Hospedales, A.~Antoniou, P.~Micaelli, and A.~Storkey, ``Meta-learning in
  neural networks: A survey,'' \emph{IEEE Transactions on Pattern Analysis and
  Machine Intelligence}, vol.~44, no.~9, pp. 5149--5179, 2022.

\bibitem{Finn2017MAML}
C.~Finn, P.~Abbeel, and S.~Levine, ``Model-agnostic meta-learning for fast
  adaptation of deep networks,'' in \emph{Proceedings of the 34th International
  Conference on Machine Learning (ICML)}, 2017, pp. 1126--1135.

\bibitem{khurana_clasp_2025}
\BIBentryALTinterwordspacing
M.~Khurana, ``Clasp (in development),'' GitHub repository (under development),
  2025, repository is currently empty; accessed: \today. [Online]. Available:
  \url{https://github.com/MKhurana07/clasp}
\BIBentrySTDinterwordspacing

\bibitem{PentestAgent_2411_05185}
\BIBentryALTinterwordspacing
X.~Shen, L.~Wang, Z.~Li, Y.~Chen, W.~Zhao, D.~Sun, J.~Wang, and W.~Ruan,
  ``Pentestagent: Incorporating llm agents to automated penetration testing,''
  2025. [Online]. Available: \url{https://arxiv.org/abs/2411.05185}
\BIBentrySTDinterwordspacing

\bibitem{OneDayAgent_2404_08144}
\BIBentryALTinterwordspacing
R.~Fang, R.~Bindu, A.~Gupta, and D.~Kang, ``Llm agents can autonomously exploit
  one-day vulnerabilities,'' 2024. [Online]. Available:
  \url{https://arxiv.org/abs/2404.08144}
\BIBentrySTDinterwordspacing

\bibitem{YURASCANNER_NDSS2025}
\BIBentryALTinterwordspacing
A.~Stafeev, T.~Recktenwald, G.~D. Stefano, S.~Khodayari, and G.~Pellegrino,
  ``Yurascanner: Leveraging llms for task-driven web app scanning,'' in
  \emph{Proceedings of the 32nd Network and Distributed System Security
  Symposium (NDSS 2025)}, San Diego, CA, USA, 2025. [Online]. Available:
  \url{https://www.ndss-symposium.org/wp-content/uploads/2025-388-paper.pdf}
\BIBentrySTDinterwordspacing

\bibitem{pentestgpt_usenix24}
\BIBentryALTinterwordspacing
Y.~Deng, H.~Dai, C.~Li, S.~Hu, X.~Zhang, S.~Ji, and H.~Wang, ``Pentestgpt:
  Evaluating and harnessing large language models for automated penetration
  testing,'' in \emph{Proceedings of the 33rd USENIX Security Symposium (USENIX
  Security '24)}, 2024. [Online]. Available:
  \url{https://www.usenix.org/system/files/usenixsecurity24-deng.pdf}
\BIBentrySTDinterwordspacing

\bibitem{zhang2023autopatch}
\BIBentryALTinterwordspacing
M.~Seo, W.~Choi, M.~You, and S.~Shin, ``Autopatch: Multi-agent framework for
  patching real-world cve vulnerabilities,'' 2025. [Online]. Available:
  \url{https://arxiv.org/abs/2505.04195}
\BIBentrySTDinterwordspacing

\bibitem{RapidPen_2502_16730}
\BIBentryALTinterwordspacing
S.~Nakatani, ``Rapidpen: Fully automated ip-to-shell penetration testing with
  llm-based agents,'' 2025. [Online]. Available:
  \url{https://arxiv.org/abs/2502.16730}
\BIBentrySTDinterwordspacing

\bibitem{hptsa}
\BIBentryALTinterwordspacing
Y.~Zhu, A.~Kellermann, A.~Gupta, P.~Li, R.~Fang, R.~Bindu, and D.~Kang, ``Teams
  of llm agents can exploit zero-day vulnerabilities,'' 2025. [Online].
  Available: \url{https://arxiv.org/abs/2406.01637}
\BIBentrySTDinterwordspacing

\bibitem{SAN2PATCH_2025}
\BIBentryALTinterwordspacing
Y.~Kim, J.~Lee, J.~Park, X.~Zheng, Y.-C.~F. Wang, H.~Ha, and T.~Xu, ``Logs in,
  patches out: Automated vulnerability repair via tree-of-thought llm
  analysis,'' in \emph{Proceedings of the 34th USENIX Security Symposium
  (USENIX Security '25)}, Seattle, WA, USA, 2025. [Online]. Available:
  \url{https://www.usenix.org/conference/usenixsecurity25/presentation/kim-youngjoon}
\BIBentrySTDinterwordspacing

\bibitem{shan2025rcacopilottransformingnetwork}
\BIBentryALTinterwordspacing
A.~Shan, J.~Kaur, R.~Singh, T.~Banka, R.~Yavatkar, and T.~Sridhar, ``Rca
  copilot: Transforming network data into actionable insights via large
  language models,'' 2025. [Online]. Available:
  \url{https://arxiv.org/abs/2507.03224}
\BIBentrySTDinterwordspacing

\bibitem{OpenRCA_ICLR2025}
\BIBentryALTinterwordspacing
J.~Xu, Q.~Zhang, Z.~Zhong, S.~He, C.~Zhang, Q.~Lin, D.~Pei, P.~He, D.~Zhang,
  and Q.~Zhang, ``Openrca: Can large language models locate the root cause of
  software failures?'' ICLR 2025 Poster on OpenReview, 2025. [Online].
  Available: \url{https://openreview.net/forum?id=M4qNIzQYpd}
\BIBentrySTDinterwordspacing

\bibitem{patchagent_usenix25}
\BIBentryALTinterwordspacing
Z.~Yu, Z.~Guo, Y.~Wu, J.~Yu, M.~Xu, D.~Mu, Y.~Chen, and X.~Xing,
  ``{PatchAgent}: A practical program repair agent mimicking human expertise,''
  in \emph{Proceedings of the 34th USENIX Security Symposium (USENIX Security
  '25)}, Seattle, WA, USA, Aug. 2025, prepublication PDF:
  \url{https://www.dataisland.org/paper/patchagent.pdf}. [Online]. Available:
  \url{https://www.usenix.org/conference/usenixsecurity25/presentation/yu-zheng}
\BIBentrySTDinterwordspacing

\bibitem{VRPilot}
\BIBentryALTinterwordspacing
U.~Kulsum, H.~Zhu, B.~Xu, and M.~d'Amorim, ``A case study of llm for automated
  vulnerability repair: Assessing impact of reasoning and patch validation
  feedback,'' 2024. [Online]. Available: \url{https://arxiv.org/abs/2405.15690}
\BIBentrySTDinterwordspacing

\bibitem{camporese2025repairingvulnerabilitiesinvisiblehands}
\BIBentryALTinterwordspacing
M.~Camporese and F.~Massacci, ``Repairing vulnerabilities without invisible
  hands. a differentiated replication study on llms,'' 2025. [Online].
  Available: \url{https://arxiv.org/abs/2507.20977}
\BIBentrySTDinterwordspacing

\bibitem{nitin2025faultlineautomatedproofofvulnerabilitygeneration}
\BIBentryALTinterwordspacing
V.~Nitin, B.~Ray, and R.~Z. Moghaddam, ``Faultline: Automated
  proof-of-vulnerability generation using llm agents,'' 2025. [Online].
  Available: \url{https://arxiv.org/abs/2507.15241}
\BIBentrySTDinterwordspacing

\bibitem{nist_sp800_61}
\BIBentryALTinterwordspacing
P.~Cichonski, T.~Millar, T.~Grance, and K.~Scarfone, ``Computer security
  incident handling guide,'' National Institute of Standards and Technology,
  Tech. Rep. NIST SP 800-61 Revision 2, 2012. [Online]. Available:
  \url{https://csrc.nist.gov/publications/detail/sp/800-61/rev-2/final}
\BIBentrySTDinterwordspacing

\bibitem{veracode_soss_2025}
\BIBentryALTinterwordspacing
{Veracode}, ``State of software security 2025,'' Veracode, Tech. Rep., 2025,
  accessed 2025. [Online]. Available:
  \url{https://www.veracode.com/wp-content/uploads/2025/02/State-of-Software-Security-2025.pdf}
\BIBentrySTDinterwordspacing

\bibitem{darpa_aixcc}
``Darpa ai cyber challenge (aixcc): Scoring and evaluation overview,''
  \url{https://aicyberchallenge.com/}, 2025, accessed Aug 25, 2025.

\bibitem{cybench_2025}
\BIBentryALTinterwordspacing
A.~K. Zhang, N.~Perry, R.~Dulepet, J.~Ji, C.~Menders, J.~W. Lin, E.~Jones,
  G.~Hussein, S.~Liu, D.~Jasper, P.~Peetathawatchai, A.~Glenn, V.~Sivashankar,
  D.~Zamoshchin, L.~Glikbarg, D.~Askaryar, M.~Yang, T.~Zhang, R.~Alluri,
  N.~Tran, R.~Sangpisit, P.~Yiorkadjis, K.~Osele, G.~Raghupathi, D.~Boneh,
  D.~E. Ho, and P.~Liang, ``Cybench: A framework for evaluating cybersecurity
  capabilities and risks of language models,'' in \emph{ICLR}, 2025. [Online].
  Available: \url{https://openreview.net/forum?id=tc90LV0yRL}
\BIBentrySTDinterwordspacing

\bibitem{autopenbench_2024}
\BIBentryALTinterwordspacing
L.~Giacchini and collaborators, ``Autopenbench: Benchmarking generative agents
  for penetration testing,'' \emph{arXiv preprint arXiv:2410.03225}, 2024.
  [Online]. Available: \url{https://arxiv.org/abs/2410.03225}
\BIBentrySTDinterwordspacing

\bibitem{VerizonDBIR2025}
\BIBentryALTinterwordspacing
VERIZON, ``2025 data breach investigations report,'' Tech. Rep., 2025.
  [Online]. Available:
  \url{https://www.verizon.com/business/resources/reports/dbir/}
\BIBentrySTDinterwordspacing

\bibitem{Ockham1347}
E.~Sober, \emph{Ockham's Razors: A User's Manual}.\hskip 1em plus 0.5em minus
  0.4em\relax Cambridge University Press, 2015.

\bibitem{LeGoues2019}
\BIBentryALTinterwordspacing
C.~L. Goues, N.~Holtschulte, E.~K. Smith, Y.~Brun, P.~Devanbu, S.~Forrest, and
  W.~Weimer, ``The manybugs and introclass benchmarks for automated repair of
  {C} programs,'' \emph{IEEE Transactions on Software Engineering}, 2015.
  [Online]. Available:
  \url{https://clairelegoues.com/assets/papers/legoues15tse.pdf}
\BIBentrySTDinterwordspacing

\bibitem{aixcc-website}
\BIBentryALTinterwordspacing
{DARPA}, ``{DARPA} {AI} cyber challenge ({AIxCC}) overview and results,'' 2025.
  [Online]. Available: \url{https://aicyberchallenge.com/overview/}
\BIBentrySTDinterwordspacing

\end{thebibliography}

\end{document}